\documentclass[a4paper,orivec,runningheads]{llncs}
\usepackage{makeidx}  
\usepackage{graphicx}
\usepackage{subfigure}
\usepackage{amsmath}
\usepackage{amssymb}
\usepackage{verbatim}
\usepackage{algorithm}
\usepackage{algorithmic}
\usepackage{booktabs}
\usepackage{multirow}
\usepackage{pbox}

\begin{document}

\title{Sequential 3D U-Nets for Biologically-Informed Brain Tumor Segmentation}
\titlerunning{Sequential 3D U-Nets for Brain Tumor Segmentation}  

\author{Andrew Beers$^1$ \and Ken Chang$^1$ \and James Brown$^1$ \and Emmett Sartor$^2$
	 \and CP Mammen$^3$ \and Elizabeth Gerstner$^{1,4}$ \and Bruce Rosen$^1$ \and Jayashree Kalpathy-Cramer$^{1,5}$}
\authorrunning{A. Beers et al.}
 
\institute{
$^1$Athinoula A. Martinos Center for Biomedical Imaging, Department of Radiology, Massachusetts General Hospital, Boston, USA\\
$^2$Massachusetts General Hospital, Boston, USA\\
$^3$NVIDIA, India\\
$^4$Athinoula A. Martinos Center for Biomedical Imaging, Department of Neuro-Oncology, Massachusetts General Hospital, Boston, USA\\
$^5$MGH \& BWH Center for Clinical Data Science, Boston, USA\\
}

\maketitle

\begin{abstract}
Deep learning has quickly become the weapon of choice for brain lesion segmentation. However, few existing algorithms pre-configure any biological context of their chosen segmentation tissues, and instead rely on the neural network's optimizer to develop such associations \textit{de novo}. We present a novel method for applying deep neural networks to the problem of glioma tissue segmentation that takes into account the structured nature of gliomas –- edematous tissue surrounding mutually-exclusive regions of enhancing and non-enhancing tumor. We trained multiple deep neural networks with a 3D U-Net architecture in a tree structure to create segmentations for edema, non-enhancing tumor, and enhancing tumor regions. Specifically, training was configured such that the whole tumor region including edema was predicted first, and its output segmentation was fed as input into separate models to predict enhancing and non-enhancing tumor. Our method was trained and evaluated on the publicly available BraTS dataset, achieving Dice scores of 0.882, 0.732, and 0.730 for whole tumor, enhancing tumor and tumor core respectively.

\end{abstract}

\section{Introduction}

Gliomas are among the most common forms of brain tumor, which are typically categorized into low-grade (LGG) and high-grade (HGG) owing to their different prognostic outcomes. Non-invasive medical imaging modalities such as computed tomography (CT) and magnetic resonance imaging (MRI) are employed to determine tumor phenotypes and inform the appropriate course of treatment, which includes chemotherapy, radiation therapy, and surgical resection. Anatomical MR sequences often include T1-weighted (T1), gadolinium-enhanced T1-weighted (T1-post), T2-weighted (T2) and fluid-attenuated inversion recovery (FLAIR). Characteristics of gliomas observed in MR images include enhancing tumor (hyperintense in T1-post compared with T1), non-enhancing/necrotic tumor (hypointense in T1-post compared with T1) and peritumoral edema (hyperintense in FLAIR). Manual delineation of these tumor regions allows for useful prognostic indicators to be extracted based on intensity, volume, shape and texture. However the annotation process requires significant effort on the part of the annotator, which can be prohibitively time consuming and subject to user error and bias. As a consequence, many research groups have developed methods to perform semi-automated or fully-automated tumor segmentation.

The Brain Tumor Segmentation (BraTS) challenge~\cite{menze2015multimodal} was established to evaluate state-of-the-art algorithms on a multi-institutional dataset of MRI scans with manually labeled tumor regions. Deep learning has become the preeminent approach to this task~\cite{kamnitsas2017efficient,havaei2017brain}, outperforming shallow classification methods across a wide variety of problem domains~\cite{lecun2015deep}. We continue this trend in our submission to BraTS 2017, which uses convolutional neural networks (CNNs) to segment tumor tissue in a biologically informed manner. Specifically, training was configured such that the whole tumor (WT) region including edema was predicted first, and its output segmentation was fed as input into separate models to predict enhancing tumor (ET) and non-enhancing/necrotic tumor core (TC). We make use of the 3D U-Net~\cite{cciccek20163d} architecture to perform segmentation in a patch-based manner, yielding encouraging results on the BraTS validation set.

\section{Methods}

\subsection{U-Net architecture and training}

\begin{figure}
	\centering
	\includegraphics[width=1.0\linewidth]{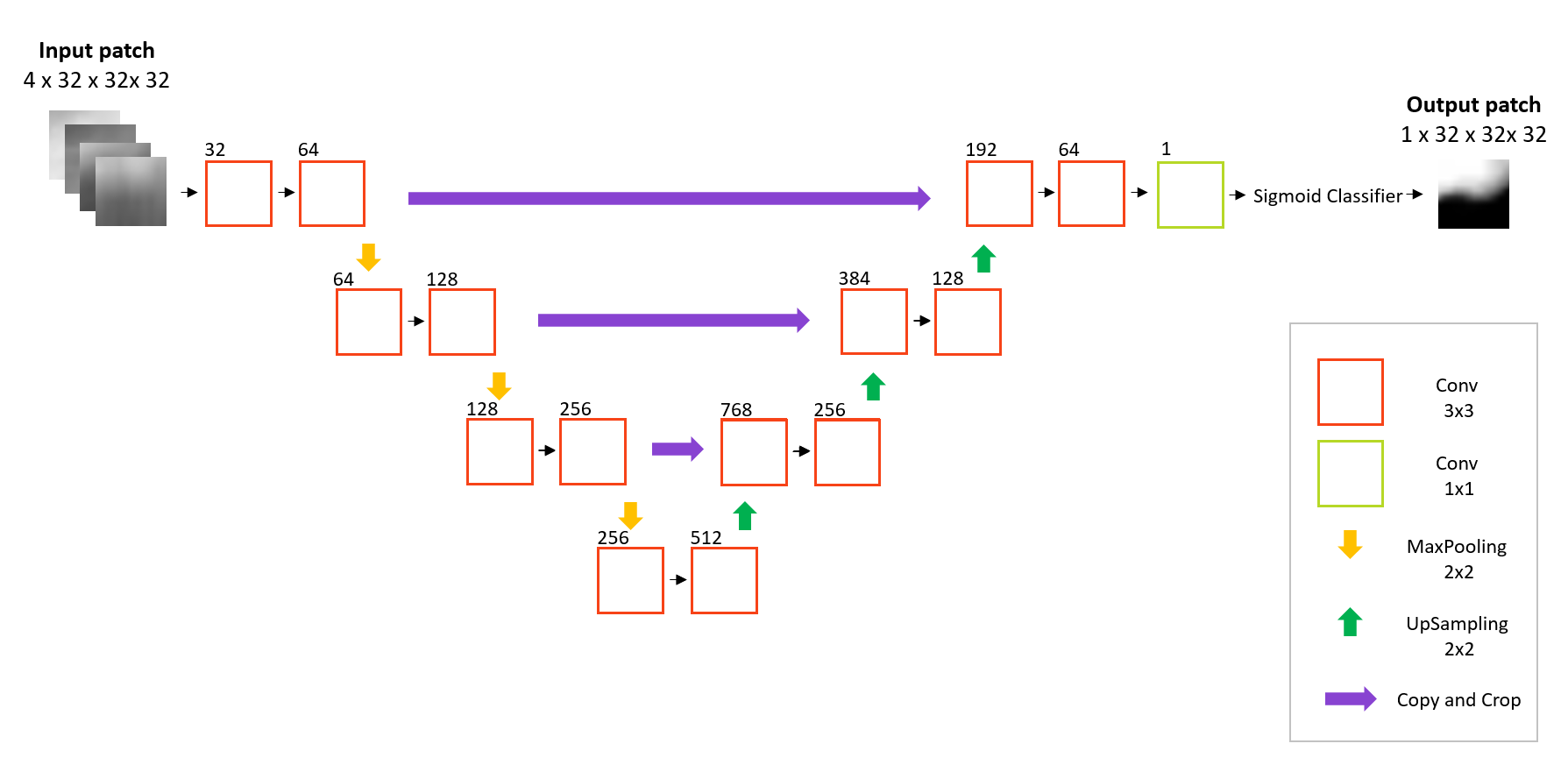}
	\caption[3D U-Net.]{The 3D U-Net architecture for glioma segmentation. 32$^3$ voxel patches are used to perform binary segmentation of each tissue type independently. The number of feature maps produced by the convolutional layers is denoted above each colored square.}
	\label{fig:unet}
\end{figure}

The U-Net architecture is shown in Fig.~\ref{fig:unet}. Similar to the original 2D U-Net \cite{ronneberger2015u}, our architecture consists of a downsampling and an upsampling arm with residual connections between the two that concatenate feature maps at different spatial scales. The networks were designed to receive input patches of size 32 $\times$ 32 $\times$ 32 voxels, comprising four MR channels; T1, T1-post, T2 and FLAIR. Rectified linear unit (ReLU) activation was used in all layers, with the exception of the final sigmoid output. Batch normalization~\cite{ioffe2015batch} was applied after each convolutional layer for regularization purposes. We used Nestorov Adaptive Moment Estimation (NAdam)~\cite{dozat2016incorporating} to train the 3D U-Nets with an initial learning rate $10^{-6}$, minimizing a soft dice loss function. Networks were trained on NVIDIA Tesla P100 GPUs for up to 200 epochs or until the validation loss plateaued. 

\subsection{Patch extraction and augmentation}

All MRI volumes (T1, T1-post, T2, FLAIR) were intensity normalized to have zero mean and unit variance. Patches were then sampled in the following ratio: 1\% background, 29\% normal brain, 70\% tumor. A total of 70 patches were extracted from each subject, and augmented by means of sagittal flips to double the training set size. At inference time, each volume is gridded into 32 $\times$ 32 $\times$ 32 patches at 16 different offsets from the upper-most corner of the image. Probability maps for each of these patches is predicted by the model, and voxels with predictions from multiple overlapping patches have their probabilities averaged. Labels are produced by binarizing the averaged probability maps at a chosen threshold.

\subsection{Segmentation pipeline}

\subsubsection{Whole tumor.}
Whole tumor (WT) is defined as the set union of the peritumoral edema, enhancing tumor (ET) and non-enhancing/necrotic tumor core (TC). To produce an initial low resolution estimate of WT, the preprocessed training data were downsampled to 2mm isotropic voxels and used to train a U-Net with four input channels. After convergence, the training data were then fed into network to yield pseudo-probabilty maps of WT at 2mm resolution. These were thresholded to produce binary labelmaps which were naively upsampled to 1mm resolution using nearest-neighbour (NN) interpolation. A second U-Net was then trained on the original 1mm anatomical MR scans, with the NN upsampled labelmap as an additional input channel.

\subsubsection{Enhancing tumor and tumor core.}
Two separate U-Nets were trained on patches comprising the four MR channels plus the WT ground truth labelmap as a fifth channel. At inference time, the WT prediction from the previous step is used in place of the ground truth. The intuition behind this is that having knowledge about the boundaries of the whole tumor will allow the CNN to become more confident in its predictions of ET and TC.

\subsubsection{Post-processing CNNs.}
Noisy segmentations are typically post-processed by means of morphological processing operations such as erosion, dilation and removal of small connected components. In this work, we instead employ two additional U-Nets that serve to tweak the predictions for ET and TC. Input patches consist of seven channels; four anatomical MR and three labelmaps corresponding to WT, ET, and TC.

\section{Experiments and Results}

\subsection{Data and implementation}

We used the multi-institutional training and validation data made available for the 2017 BraTS challenge\footnote{http://http://braintumorsegmentation.org/}~\cite{menze2015multimodal}. A set of pre-operative scans are provided with T1, T1-post, T2 and FLAIR volumes for each subject.  Data were already intra-subject registered and skull-stripped. The full training set (n = 285) was randomly split into a smaller training (n = 258) and validation (n = 29) set for training and hyperparameter optimization, respectively. We report Dice scores, sensitivities, specificities for the BraTS validation set (n = 46). Ground truth labelmaps are not made available for these data, and so scores were determined by uploading our solutions to the BraTS online portal. Results for the testing data were not available at the time of publication. Our networks were implemented in Keras\footnote{https://keras.io/}, with a TensorFlow\footnote{https://www.tensorflow.org/} backend.

\subsection{Segmentation results}

\begin{figure}[t]
	\centering
	\includegraphics[width=.9\linewidth]{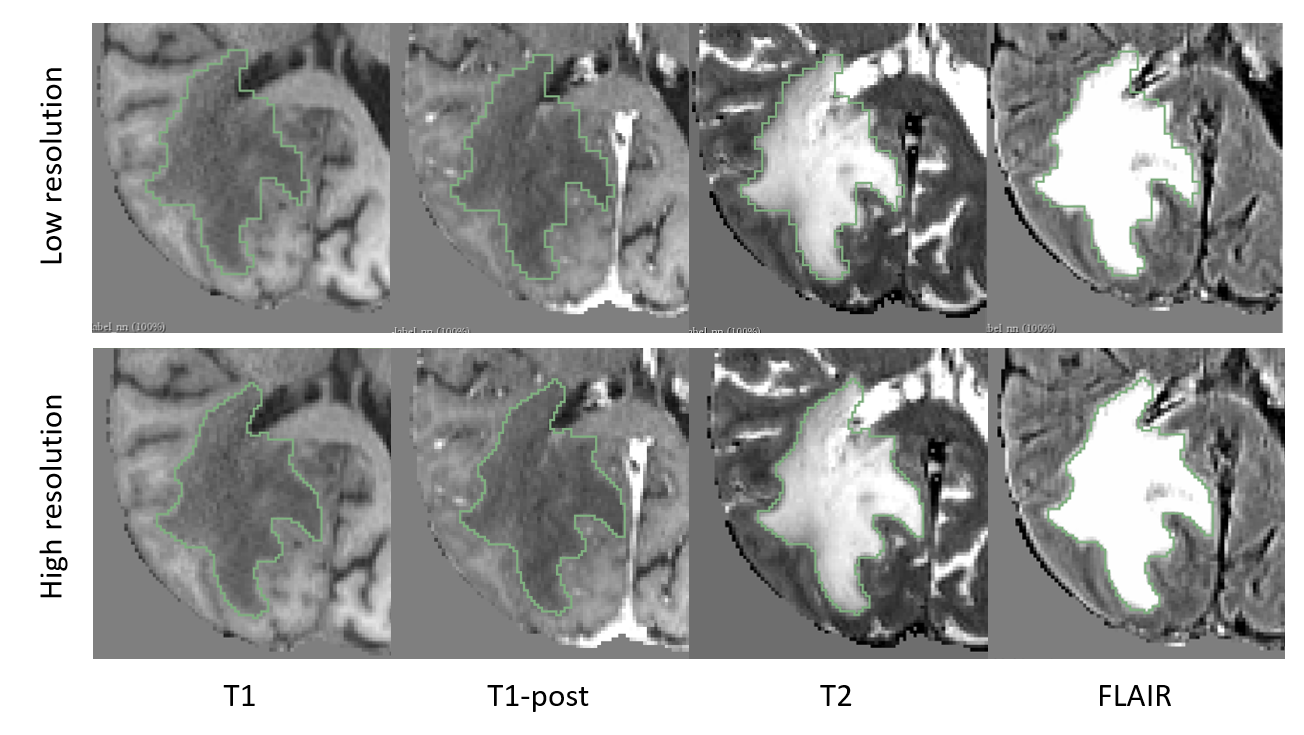}
	\caption[Whole tumor example]{Example segmentation result for whole tumor. Low resolution (top) labelmaps are upsampled using nearest-neighbour interpolation, and upsampled to high resolution (bottom).}
	\label{fig:wholetumor}
\end{figure}

\begin{figure}
	\centering
	\includegraphics[width=.9\linewidth]{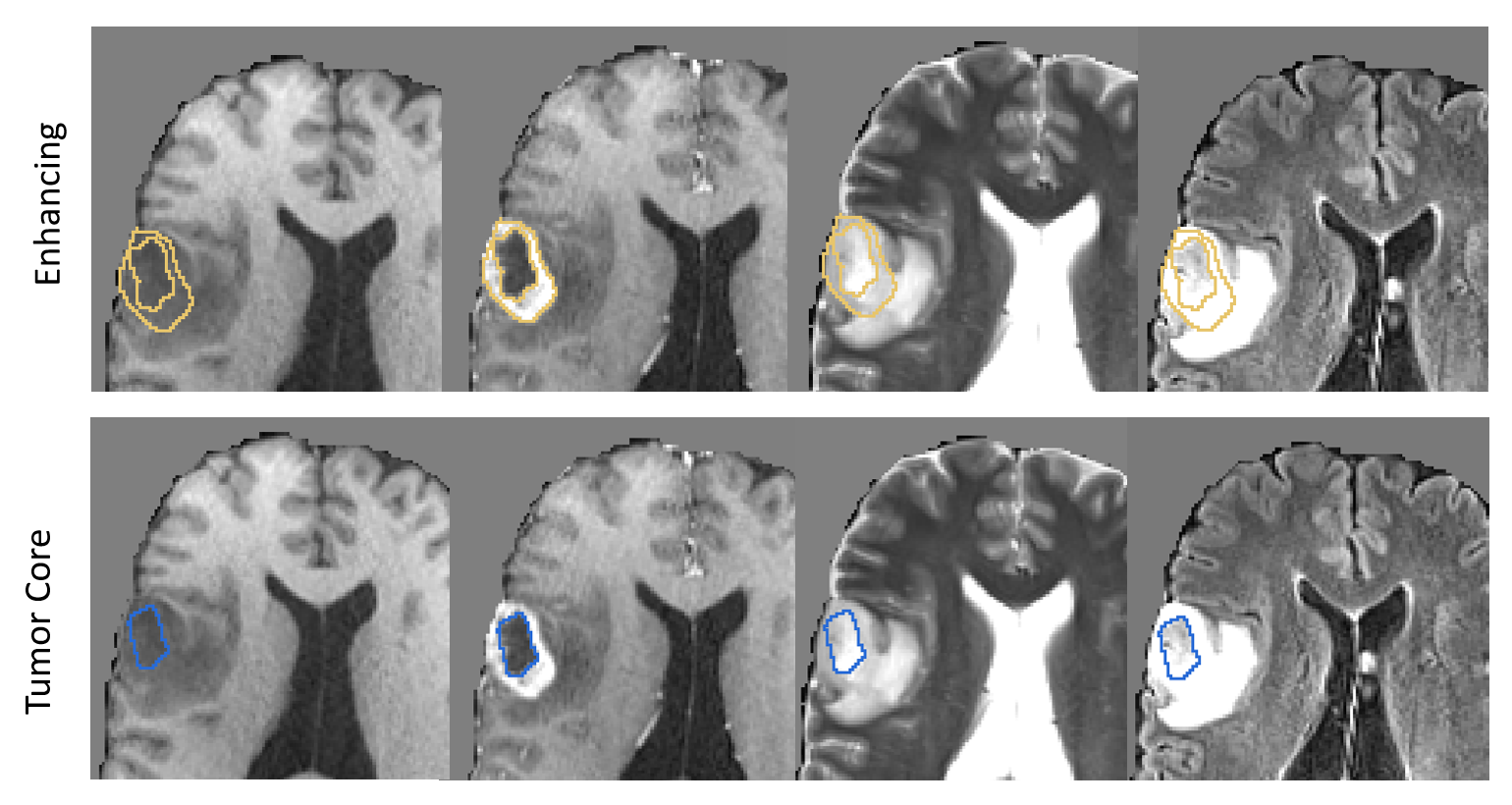}
	\caption[Whole tumor example]{Example segmentation results for enhancing tumor and tumor core.}
	\label{fig:et_tc}
\end{figure}

\begin{figure}
	\centering
	\includegraphics[width=\linewidth]{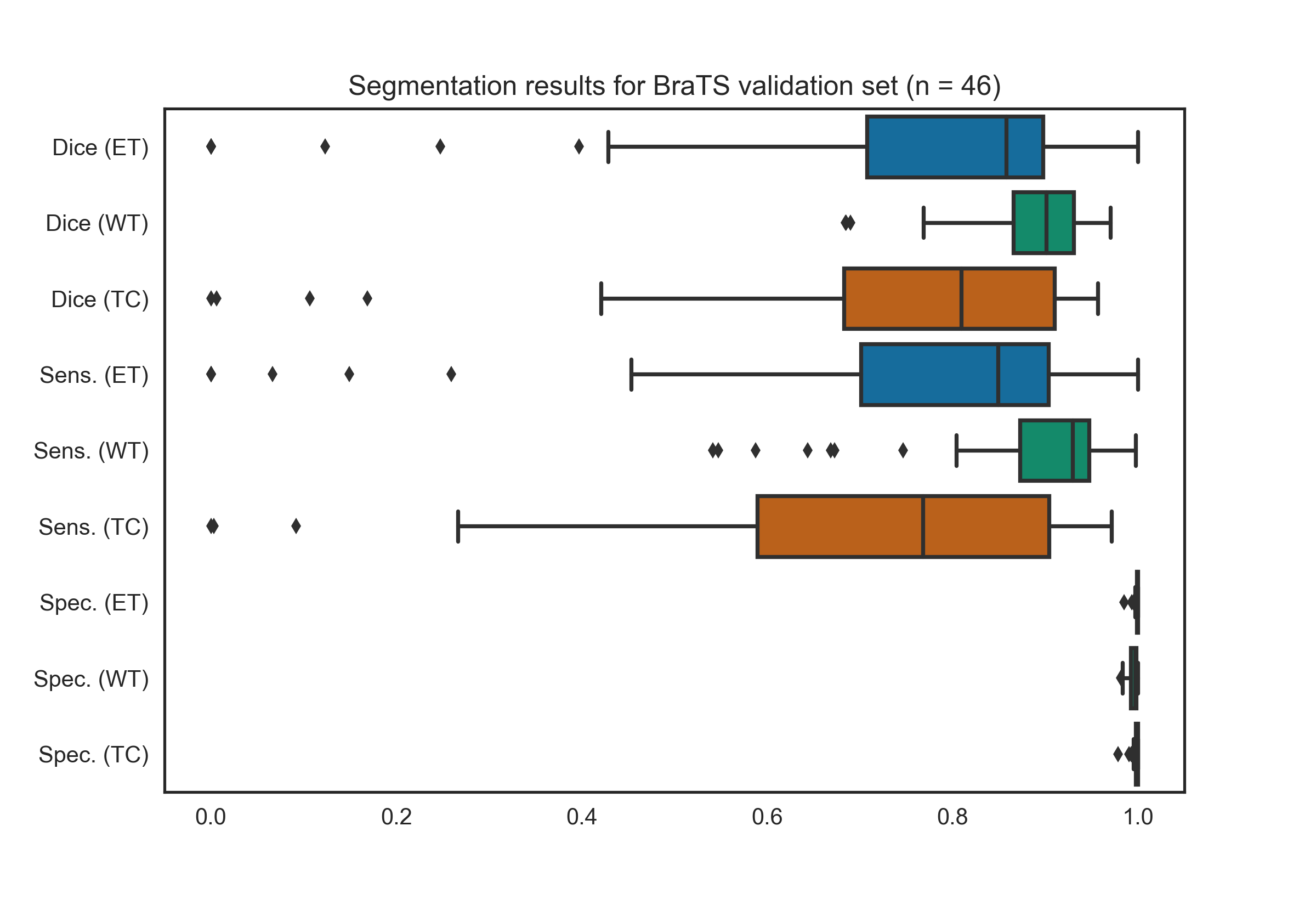}
	\caption[Validation results]{Box plot showing segmentation results for the BraTS validation set (n = 46). Dice scores, sensitivities and specificities are shown for enhancing tumor (ET), whole tumor (WT) and tumor core (TC).}
	\label{fig:boxplot}
\end{figure}

Figures~\ref{fig:wholetumor} and~\ref{fig:et_tc} shows example segmentation results for whole tumor, enhancing tumor and tumor core. Results over the BraTS validation set are shown in Figure~\ref{fig:boxplot}. These results reflect those reported via the online leaderboard on 8th September 2017~\footnote{https://www.cbica.upenn.edu/BraTS17/lboardValidation.html}. Out of the 61 entrants, our algorithm places 21st for whole tumor, 16th for enhancing tumor and 28th for tumor core. 

\section{Conclusion}
A method based on sequential U-Nets for segmentation of glioma from multi-modal MRI is presented. We achieved mean dice scores over the BraTS 2017 validation set (n = 46) of 0.882, 0.732, and 0.730 for whole tumor, enhancing tumor and tumor core respectively.

\subsubsection{Acknowledgements.}\label{sec:acknowledgements}
This research was carried out in whole or in part at the Athinoula A. Martinos Center for Biomedical Imaging at the Massachusetts General Hospital, using resources provided by the Center for Functional Neuroimaging Technologies, P41EB015896, a P41 Biotechnology Resource Grant supported by the National Institute of Biomedical Imaging and Bioengineering (NIBIB), National Institutes of Health. This work was supported by a training grant from the NIH Blueprint for Neuroscience Research (T90DA022759/R90DA023427), and the NCI/NIH (U24CA180927, U01CA154601). We would also like the acknowledge the GPU computing resources provided by the MGH and BWH Center for Clinical Data Science.
%
%
\bibliographystyle{unsrt}


%
%

\end{document}